\documentclass[letterpaper, 10 pt, conference]{ieeeconf}  

\IEEEoverridecommandlockouts                              
\usepackage{balance}
\usepackage{subfiles}
\usepackage{algorithmic}
\usepackage{mathbbol}
\usepackage{siunitx}

\usepackage{amsmath,amssymb,mathrsfs,bm}
\usepackage[ruled, linesnumbered, vlined]{algorithm2e}
\usepackage{comment}
\usepackage{graphicx}
\usepackage{graphics}
\usepackage[tight]{subfigure}
\usepackage{wrapfig}
\usepackage{float}
\usepackage[table]{xcolor}
\usepackage{multirow}
\usepackage{array}
\usepackage{enumerate}
\usepackage{sidecap}
\usepackage{flushend}
\usepackage{hyperref}
\usepackage{url}
\graphicspath{{./figs/}{./}}

\usepackage[space]{cite}

\usepackage{color}

\newcommand{\vect}[1]{\mathbf{#1}}

\definecolor{lightblue}{rgb}{0,0.2,1}
\definecolor{black}{rgb}{0,0,0}
\hypersetup{
    colorlinks=true,%
    citebordercolor=white,%
    filebordercolor=white,%
    linkbordercolor=white,%
    urlbordercolor=white,%
    citecolor=black,%
    linkcolor=black,%
    urlcolor=black%
}

\newcounter{tecounter}
\title{\LARGE \bf Learning Partially Structured Environmental Dynamics for Marine Robotic Navigation}

\author{Chen Huang$^{1}$, Kai Yin$^{2}$, Lantao Liu$^{3}$
\thanks{$^{1}$Chen Huang is with the Department of Computer Science at the University of Southern California, Los Angeles, CA 90089, USA. E-mail:
        {\tt\small huan574@usc.edu}}%
\thanks{$^{2}$Kai Yin is with HomeAway, Inc., Austin, TX, USA. E-mail:
        {\tt\small yinkai1000@gmail.com }}%
\thanks{$^{3}$Lantao Liu is with the Department of Intelligent Systems Engineering at  Indiana University - Bloomington,
        Bloomington, IN 47408, USA. E-mail:
        {\tt\small lantao@iu.edu}}%
\thanks{A preliminary version of this work appeared
as a poster in AAAI 2018 Spring Symposium on Integrating Representation, Reasoning, Learning, and Execution for Goal Directed Autonomy.}
}

\begin{document}

\maketitle
\thispagestyle{empty}
\pagestyle{empty}



\begin{abstract}
We investigate the scenario that a robot needs to reach a designated goal after taking a sequence of appropriate actions in a non-static environment that is partially structured.
One application example is to control a marine vehicle to move in the ocean. 
The ocean environment is dynamic and oftentimes the ocean waves result in strong disturbances that can disturb the vehicle's motion. 
Modeling such dynamic environment is non-trivial, and integrating such model in the robotic motion control is particularly difficult. Fortunately, the ocean currents usually form some local patterns (e.g. vortex) and thus the environment is partially structured. 
The historically observed data can be used to train the robot to learn to interact with the ocean tidal  disturbances.  
In  this  paper  we  propose a  method  that applies the deep reinforcement learning framework to learn such partially structured  complex disturbances.
Our results show that, by training the robot under artificial and real ocean disturbances, the robot is able to successfully act in complex and spatiotemporal environments.
\end{abstract}


\section{Introduction and Related Work}

Acting in unstructured environments can be challenging especially when the environment is dynamic and involves continuous control states.
We study the goal-directed action decision-making problem where a robot's action can be disturbed  by  environmental  disturbances  such  as  the  ocean waves or air turbulence.

To be more concrete, consider a scenario where an underwater vehicle navigates across an area of ocean over a period of a few weeks to reach a goal location. 
Underwater vehicles such as autonomous gliders currently in use can travel long distances but move at speeds comparable to or slower than, typical ocean currents~\cite{Wynn2014451,Smith2011a}. 
Moreover, the disturbances caused by ocean eddies oftentimes are complex to be modeled.
This is because when we navigate the underwater (or generically aquatic) vehicles, we usually consider long term and long distance missions, and during this process the ocean currents can change significantly, causing spatially and temporally varying disturbances.
The ocean currents are not only complex in patterns, but are also strong in tidal forces and can easily perturb the underwater vehicle' motion, causing significantly uncertain action outcomes.

In general, such non-static and diverse disturbances are a reflection of the unstructured natural environment, 
and oftentimes it is very difficult to accurately formulate the complex disturbance dynamics using mathematical models.  
Fortunately, many disturbances caused by nature are seasonal, recurring, and can be observed, and the observation data is available for some time horizons. 
For example, we can get the forecast, nowcast, and hindcast of the weather including the wind (air turbulence) information from related observatories. 
Similarly, the ocean currents information can also be obtained,  and using such data allows us to train the robot to learn to interact with the ocean currents which is spatiotemporal.

Learning spatiotemporal features have been well investigated. For instance, 
the latent spatiotemporal features between two images can be retrieved through convolutional and recurrent learning frameworks~\cite{taylor2010convolutional,tran2015learning,jain2016structural}. Unfortunately, most existing spatiotemporal deep learning algorithms do not involve agent decision-making mechanisms. 

\begin{figure} [t]
  \centering
        {\label{fig:currents}\includegraphics[width=3.2in]{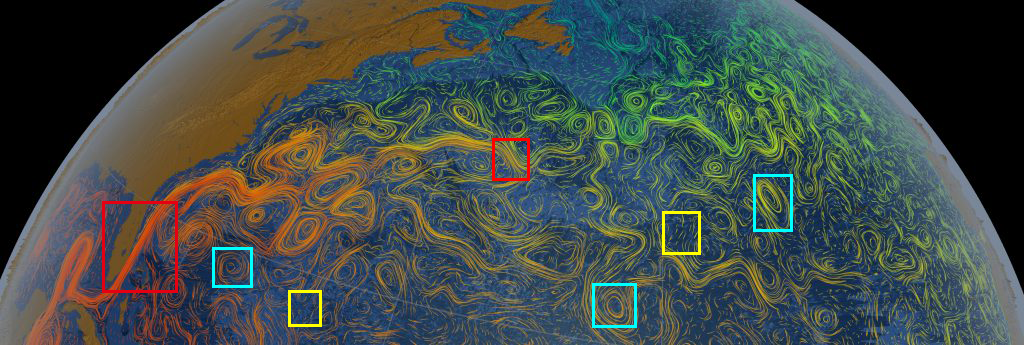}}
  \caption{Ocean currents consist of local patterns (source: NASA). Red box: uniform pattern. Blue box: vortex. Yellow box: meandering}
\label{fig:ocean-currents} \vspace{-10pt}
\end{figure}

Recently, studies on deep reinforcement learning have revealed a great potential for addressing complex decision problems such as robotic control~\cite{levine2016end, Guetal17}, computer and board game playing~\cite{mnih2015human,silver2017mastering,dqn-beer}. 
We found that there are certain similarities between our marine robots decision-making and the game playing scenarios if one regards the agent's interacting platform/environment here is the nature instead of a game.
However, one critical challenge that prevents robots from using deep learning is the lack of sufficient training data~\cite{pinto2016supersizing}. Indeed, using robots to collect training data can be extremely costly (e.g., in order to get one set of marine data using onboard sensors, it is not uncommon that a marine vehicle needs to take a few days and traverse hundreds of miles).
Also, modeling a vast area of environment can be computationally expensive.

Fortunately, the complex-patterned disturbance usually can be characterized by local patches, where a single patch may possess a particular disturbance pattern (e.g., a vortex/ring pattern, or a river-flow-like meandering pattern, or a constant waving pattern)~\cite{oey2005loop}, and the total number of the basic patterns are enumerable.
Therefore, we are motivated by training the vehicle to learn those local patches/patterns offline so that during the real-time mission, if the disturbance is a mixture of a subset of those learned patterns, the vehicle can take advantage of what it has learned to cope with it easily, thus reducing the computation time for online action prediction and control.

We use the iterative linear quadratic regulator~\cite{li2004iterative} to model the vehicle dynamics and control, and use the policy gradient framework~\cite{levine2013guided} to train the network.
We tested our method on simulations with both artificially created dynamic disturbances as well as from a history of ocean current data, and our extensive evaluation results show that the trained robot achieved satisfying performance.

\section{Technical Approach}

We use the deep reinforcement learning framework to model our decision-making problem.
Specifically, we use $s$ and $a$ to denote the robot's state and action, respectively. 
The input of the deep network is the disturbance information which is typically a vector field. Our goal is to obtain a stochastic form of policy $\pi_{\eta}(s, a) = \text{P}(a | s, \eta)$ parameterized by $\eta$ that maximizes the
discounted, cumulative reward $R_{t} = \sum_{t'=t}^{T} \gamma^{t'-t} r_{t'}$, 
where $T$ is a horizon term specifying the maximum time steps and $r_t$ is the reward at time $t$ and $\gamma$ is a discounting constant between 0 and 1 that ensures the convergence of sums. 
In this study, a deep recurrent neural network is used to
approximate the optimal action-value function $Q^*(s, a) = \displaystyle\max_{\pi} \text{E}[R_{t} | s_{t}, a_{t}, \pi] $. Therefore, the policy parameters $\eta$ are the weights of the neural network.
More details of the basic model can be found in~\cite{mnih2015human}.

\subsection{Network Design}

Since the ocean currents data over a period is available,  
we build our neural network with an input that integrates both the ocean (environmental) and the vehicle's states.
The environmental state here is a vector field representing the ocean currents (their strengths and directions).

\begin{figure}[t]
  \centering
  \includegraphics[width=0.5\textwidth]{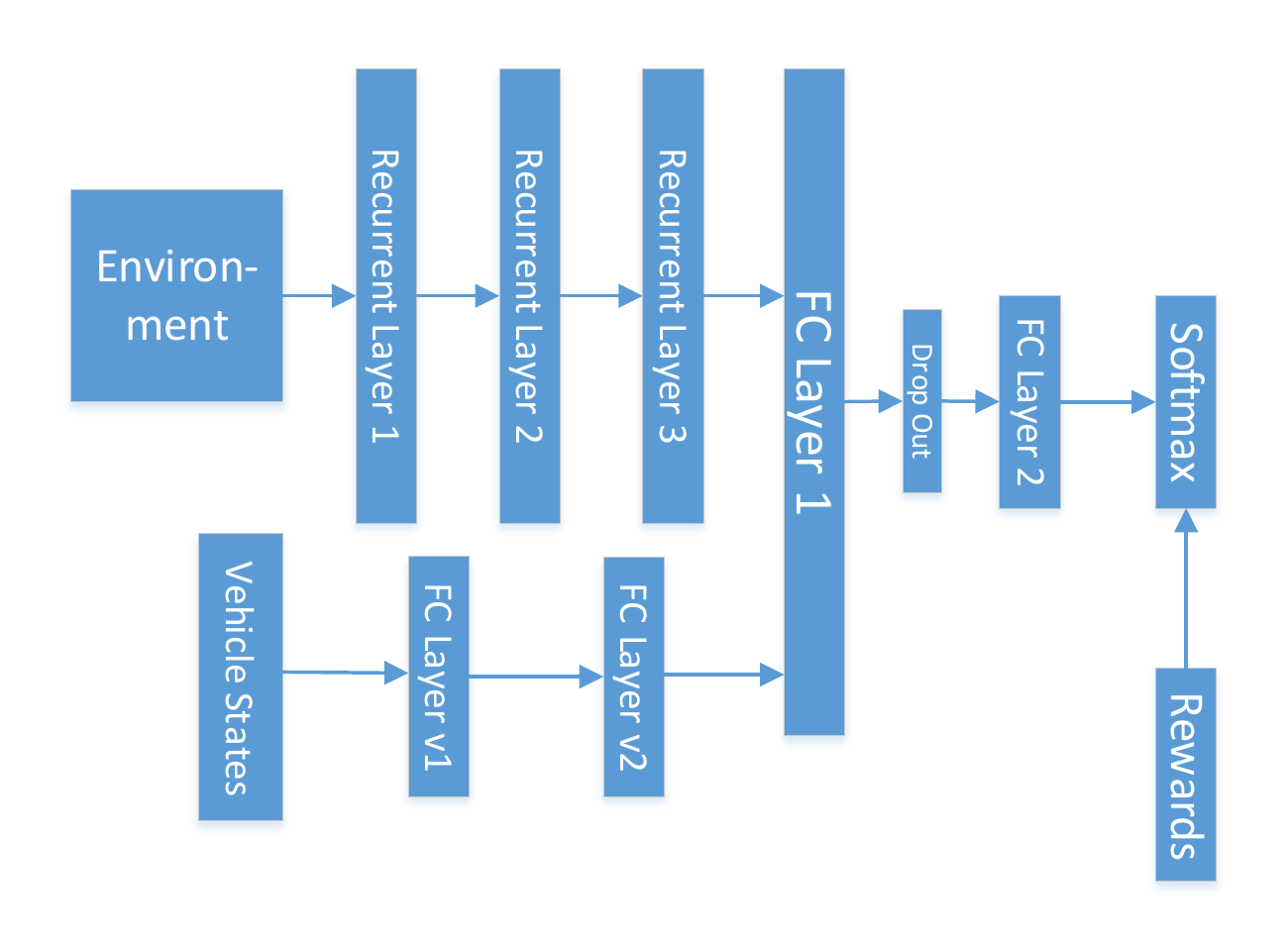}
  \caption{Neural network structure}
  \label{fig:network}
\end{figure}

The structure of the neural network is shown in Fig.~\ref{fig:network}. 
Specifically, the input consists of two components: environment and vehicle states. The environmental component has three channels, where the first two channels convey the information about the $x$-axis and $y$-axis of the disturbance vector field, and the third describes the goal and obstacles.
We assume that each grid of the input map in the third channel has three types: it can be occupied by obstacle  (we set its value -1), or be free/empty for robot to transit to (with value 0), or be occupied by the robot (with value 1). 
The vehicle state component of the input is a vector that includes the vehicle's velocity and its direction towards the goal.
Note that we do not include the robot's position in input because we want the robot to be sensitive only to environmental dynamics but not to specific (static) locations. 

The design of internal hidden layers is depicted in Fig.~\ref{fig:network}. 
The front 3 recurrent layers process the environment information, while the vehicle states begin to be combined from the first fully connected (FC) layer. 
The reason of such a design lies in that the whole net could be regarded as two correlated and eventually connected sub-nets: one sub-net is used to characterize features of disturbances, which is analogous to that of image classification; the other sub-net is a decision component for choosing the best action strategy. 
In addition, such separation of the inputs can reduce the number of parameters so that the training process can be accelerated.

\begin{figure}[t] 
  \centering
  \includegraphics[width=0.5\textwidth]{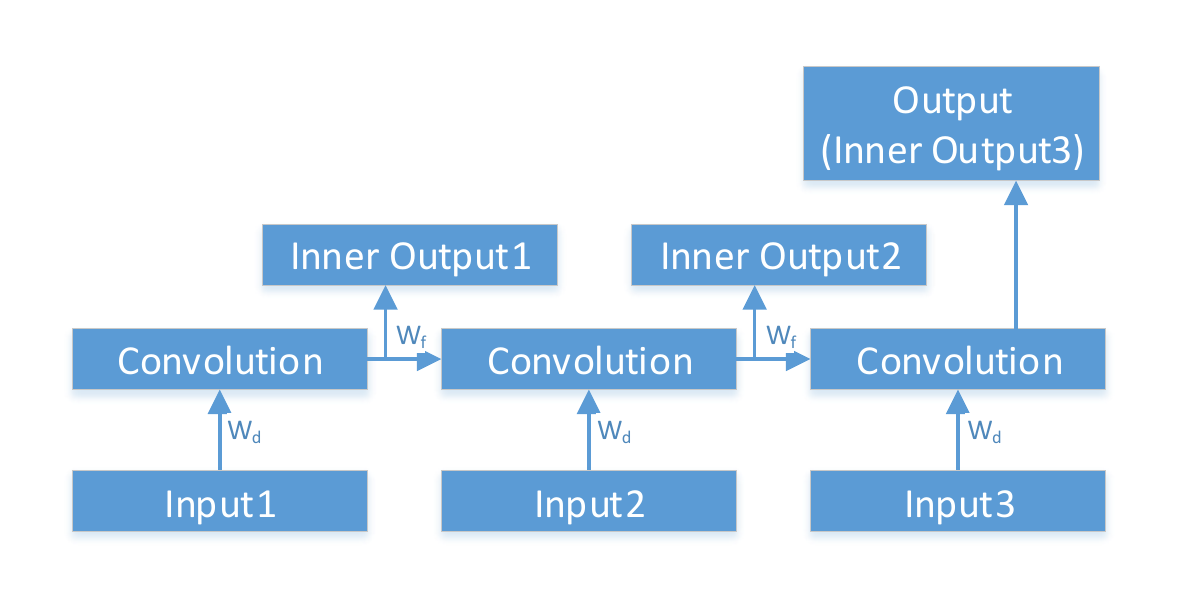}
  \caption{The structure of each recurrent layer: 
  $W_d$ stands for the weight matrix for input data, and $W_f$ represents the weight matrix for feedback data. All inner outputs are used by the next recurrent layer, and 
  only the final inner output (inner output 3) of the last recurrent layer is exported to the next FC layer.}
  \label{fig:rnn_layer}
\end{figure}

The structure for each recurrent layer is depicted in Fig.~\ref{fig:rnn_layer}. For each recurrent layer, we unfold it into 3 time steps and therefore a sequential input consisting of 3 time steps is required by each layer. Each recurrent layer will generate 3 outputs, which will be used by the following recurrent layer. 
Note that, for the last recurrent layer (i.e., layer 3), only the last output (output generated at time step 3) will be passed to the FC layer for further process. After each recurrent layer a max-pool is applied. 
The vehicle states first pass through 2 FC layers in its sub-net, and then are combined with the environmental component output from recurrent layer 3 as the input to a successive FC Layer 1. Between FC Layer 1 and 2 there exists a drop-out layer to avoid overfitting. The Softmax layer is used to normalize outputs for generating a probability distribution that can be used for sampling future actions. 
Additionally, the {\em loss funciton} is calculated using this probability distribution as well as the actual rewards. 

\subsection{Motion with Environmental Dynamics}
\label{sec:dynamics}

We consider robot's motion on a two-dimensional ocean surface, and assume the robot's state $ z = (x, y, \theta)$, which includes the vehicle's position $(x, y)$ and orientation $\theta$ in the world frame, respectively.
Since the behavior of the vehicle in the 2D environment is similar to that of the ground mobile robot, thus we opt to use a Dubins car model to simulate its motion. Similar settings can also be found in~\cite{mahmoudian2008underwater, manzanilla2018design,goerzen2010survey,Smith2011a,hvamb2015motion}. 
The dynamics can be written as:
\begin{equation}
    \dot{x} = v \cos\theta, \quad \dot{y} = v \sin \theta, \quad \dot{\theta} = u ,
\end{equation}
where $v$ and $u$ are the vehicle's linear and angular speeds, respectively. 


One challenge here is that, the vehicle dynamics is not only nonlinear, but also time varying due to the non-negligible spatiotemporal ocean disturbances $\pmb{\omega}_t(x, y)=[\omega_x (x,y), \omega_y (x,y)]$. Here $\pmb{\omega}_t(x, y)$ is assumed to be a deterministic function with differentiable components, and no explicit notation $t$ is used for $\omega_x (x,y)$ and $\omega_y (x,y)$ for simplicity. Since the ocean environment is usually open with little or no obstacles on or near the surface, consequently we assume that the robot's linear speed $v$ is slowly time-varying and can be viewed as a constant during the planning horizon, and assume the control input is angular speed $u$. Therefore, the motion dynamics that integrate external time-varying disturbance becomes
\begin{align}
\left\{
\begin{aligned}
\dot{x} &= v\cos{\theta} + \omega_x (x,y) \\
\dot{y} &= v\sin{\theta} + \omega_y (x,y) \\
\dot{\theta} &= u
\end{aligned}
\right. \label{eqn:time-varying-dynamics}
\end{align}
where $v$ is the magnitude of the robot velocity $\vect{v} = [v_x, v_y]^T = [v\cos{\theta}, v\sin{\theta}]$, $u$ is the control variable. The magnitude of ocean disturbances is assumed to be similar to or smaller than that of the robot velocity such that the robot can reach the destination through the control policy.


Such nonlinear control problem can be solved using the iterative Linear Quadratic Regulator (iLQR)~\cite{li2004iterative,tassa2014control}. 
However, similar to LQR, the iLQR basically assumes that the system dynamics are known so that the solution to nonlinear case can be approximated and converged. In our case, the motion dynamics Eq.~\eqref{eqn:time-varying-dynamics} largely depends on the external disturbance which is time-varying and also hard to accurately model. In our work, we approximate such complex spatiotemporal dynamics and assume that in the near-future horizon and around the local area (patch), the external disturbance is known (approximatible and predictable). 
Specific formulations can be found in Appendix. We also compared our deep learning framework with the baseline iLQR strategy in the experiments.

\subsection{Loss Function and Reward}

We employ the policy gradient framework for solution.
With the stochastic policy $\pi_{\eta} (s, a)$ and the Q-value $Q_{\pi_\eta}(s, a)$ for the state-action pair where $\eta$ denotes the parameters to be learned, the policy gradient of loss function $L(\eta)$ can be defined as follows:
\begin{equation}
\nabla_\eta L(\eta) = \mathbb{E}_{\pi_{\eta}} \big[ Q_{\pi_\eta}(s, a)  \nabla_{\eta} \text{log} \pi_\eta(s, a) \big].
\end{equation}

To improve the sampling efficiency and accelerate the convergence, we adopt the {\em importance sampling} strategy using guided samples~\cite{levine2013guided}.

With the objective of reaching the designated goal, our rewarding mechanism is related to minimize the cost from start to goal. 
The main idea is to reinforce with a large positive value for those correct actions that lead to reaching the goal quickly, and to penalize those undesired actions (e.g., those take long time or even fail to reach the goal) with small positive or negative values. Formally, we define the reward $r$ of each trial/episode as:
\begin{equation}
	\label{eq:reward}
	r = 
		\begin{cases}
			r_s, \quad &\text{succeeded}, \\
			-(\alpha r_s + (1-\alpha) r_d), \quad &\text{failed}, 
		\end{cases}
\end{equation}
where 
\begin{align}
    r_s &= \frac{1}{\sum_t{ \pi_{\eta}(s, a) ||p_t - p_G ||_2}}\label{eq:reward_guided},\\
    r_d &= 1 - e^{-D_{min}}\label{eq:penalty_guided},
\end{align}
where $||p_t - p_G ||_2$ denotes the distance from the $t$-th step position to the goal, and $D_{min} = \text{min}_t ||p_t - p_G ||_2$ is the minimum of such distance along the whole path 
(if the robot fails to reach the goal, $D_{min}$ is a non-zero value). 
The term $r_s$ in Eq.~\eqref{eq:reward_guided} evaluates the state with respect to the goal state, whereas the term $r_d$ in Eq.~\eqref{eq:penalty_guided} summarizes an evaluation over the entire path. 
Coefficient $\alpha\in[0, 1]$ is an empirical value to scale between $r_s$ and $r_d$ so that they contribute about the same to the total reward $r$. 

\subsection{Offline Training and Online Decision-Making}

We train the robot by setting different starting and goal positions in the disturbance field, and the {\em experience replay}~\cite{mnih2015human, riedmiller2005neural} mechanism is employed to avoid over-fitting.
Specifically, we define an {\em experience} as a 3-tuple $(s, a, r)$ consisting of a sequence $s$ of three consecutive states (i.e., the current state and two prior states), an action $a$, and a reward $r$. The idea is to store those experiences obtained in the past into a dataset.
Then during the reinforcement learning update process, a mini-batch of experiences is sampled from the dataset each time for training. The process of training is described in Algorithm~\ref{alg:training}, which can be summarized into four steps.
\begin{enumerate}
\item Following the current (learned) action policies, sample actions and finish a trial path or an episode.
\item Upon completion of each episode, obtain corresponding rewards (a list) according to whether the goal is reached, and assign rewards to the actions taken on that path. 
\item Add all these experiences into dataset. If the dataset has exceeded the maximum limit, erase as many as the oldest ones to satisfy the capacity.
\item Sample a mini-batch of experiences from the dataset. This batch should include the most recent path. Then shuffle this batch of data and feed them into the neural network for training. If current round number is less than the max training rounds, go back to step 1.
\end{enumerate}

\begin{algorithm}
\caption{Training}
\label{alg:training}
\begin{algorithmic}
	\STATE $round \gets 0$
	\WHILE {$round < n$}
		\STATE Obtain $List \langle s,a \rangle$ of this episode. 
		\STATE $experiences \gets \varnothing$
		\FORALL {$\langle s,a \rangle \in List \langle s,a \rangle$}
			\STATE $r \gets get\_reward(s,a)$
			\STATE $experiences \gets experiences \bigcup \langle s,a,r \rangle$
		\ENDFOR
		\STATE $subset \gets experiences$
		\STATE pad up $subset$ to batch size with data from dataset
		\STATE store $experiences$ into dataset
		\STATE shuffle $subset$
		\STATE feed $subset$ into neural network
		\STATE perform back propagation
		\STATE $round \gets round + 1$
	\ENDWHILE
\end{algorithmic}
\end{algorithm}



With the offline trained results, the decision-making is straightforward: only one forward propagation of the network with small computational effort is needed. This also allows us to handle continuous motion and unknown states.



\newcommand{\tabincell}[2]{\begin{tabular}{@{}#1@{}}#2\end{tabular}}

\section{Results}

We validated the method in the scenario of marine robot goal-driven decision making, where the ocean disturbances vary both spatially and temporally.
An underwater glider simulator written in C++ was built in order to test the proposed approach. We assume the glider glides near the surface and the ocean currents do not vary within a small depth under the surface area.
The underwater glider is modeled with simplified dynamics described in Section~\ref{sec:dynamics}. 
Thus, the simulation environment was constructed as a two-dimensional ocean surface,
and the spatiotemporal ocean currents are external disturbances for the robot and are represented as a vector field, with each vector representing the water flow velocity captured at a specific moment in a specific location. 
Specifically, each disturbance vector at location $(x, y)$ and time $t$ is denoted as $\pmb{\omega}_t(x, y) = [\omega_x (x,y), \omega_y (x,y)]$, where vector $\omega_x (x,y)$ denotes the {\em easting} velocity component (along latitude axis) and vector $\omega_y (x,y)$ denotes the {\em northing} component (along longitude axis). 
The simulator is able to read and process ocean current data from the 
Regional Ocean Model System (ROMS)~\cite{shchepetkin_regional_2005}, where the ocean current data is labelled with latitude $lat$ (corresponding to $x$), longitude $lon$ (corresponding to $y$), the current easting (corresponding to $w_x(x, y)$) and northing components (corresponding to $w_y(x, y)$), as well the time stamp. The collected historical ocean data is used to train our agent.

\begin{figure}[htbp]
  \centering
  \includegraphics[width=0.8\columnwidth]{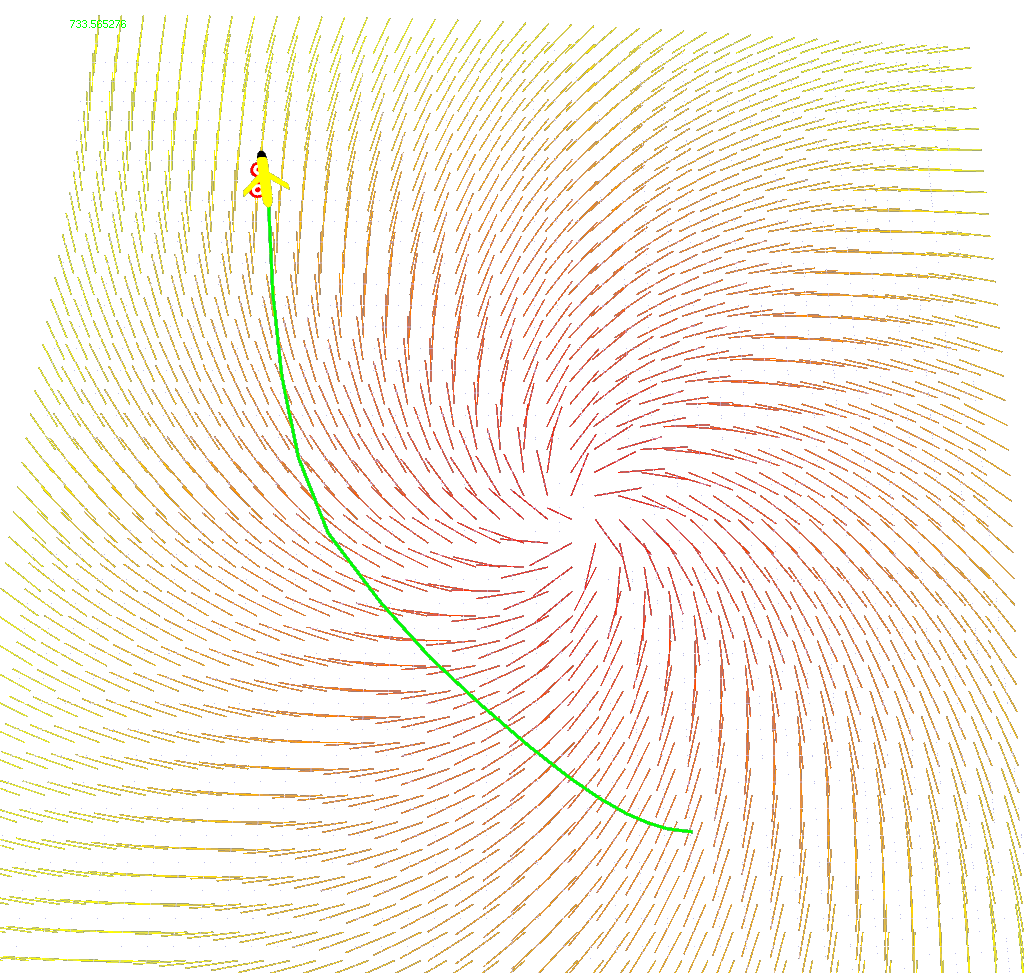}
  \caption{Demonstration of a learned path under artificially generated 
  disturbances. The center of this vortex-like vector field is translating back and forth along the diagonal direction of the simulation environment. Color represents strength of the disturbance. }
  \label{fig:arti_demo}
\end{figure}

\begin{figure}[htbp]
  \centering
  \subfigure[Input]
  {
    \includegraphics[width=0.6\columnwidth]{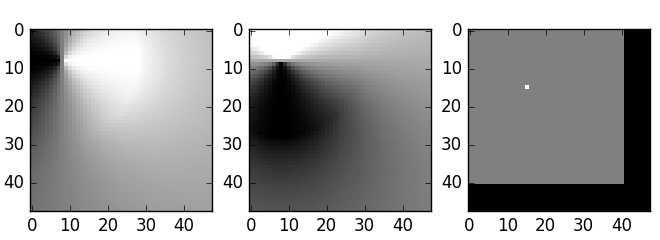}
    \label{fig:input}
  }
  \subfigure[Mix-input]
  {
    \includegraphics[width=0.2\columnwidth]{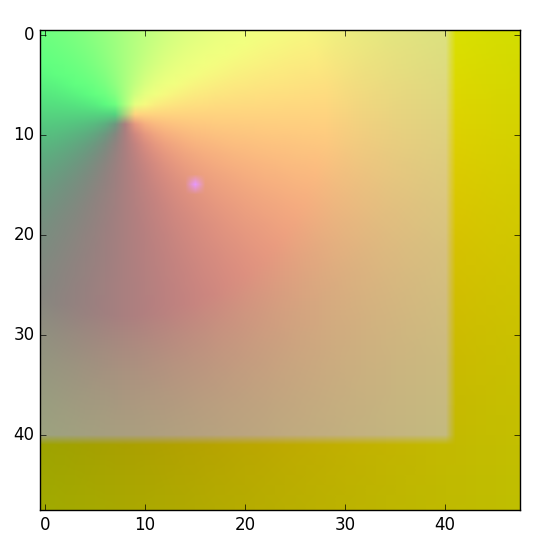}
    \label{fig:input_mix}
  }
  \subfigure[Recurrent Layer 3]
  {
    \includegraphics[width=1.\columnwidth]{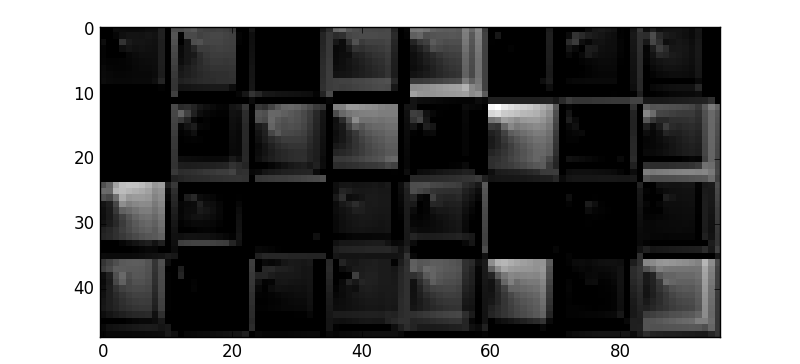}
    \label{fig:conv3}
  }
  \caption{Illustration of disturbance features captured by hidden layers}
  \label{fig:feature_map}
\end{figure}

\subsection{Network Training}

We use Tensorflow~\cite{tensorflow2015-whitepaper} to build and train the network described in Fig.~\ref{fig:network}. 
In our experiments, the input vector field map is $48\times48$, and the size of dataset for action replay is set to 10000. The learning rate is $\num{1e-6}$, 
the coefficient $\alpha$ of Eq.~\eqref{eq:reward} is set to $0.9$, 
and the batch size used for each iteration is 500. In addition,
we set the length of each episode as 300 steps.





Fig.~\ref{fig:arti_demo} demonstrates a path learned by the network under a vortex-like  disturbance field, and Fig.~\ref{fig:feature_map} shows the features extracted from internal layers of the network at some point on that path.
Fig.~\ref{fig:input} illustrates the feature of the disturbance vector field. Specifically, the first two channels of Fig.~\ref{fig:input}  are $x$ and $y$ components of the vector field, and the grey-scale color represents the strength of disturbance. 
The third channel of Fig.~\ref{fig:input} is a pixel map that contains the goal point (white dot) and obstacle information (black borders). 
Other grey grids denote free place. 
Fig.~\ref{fig:input_mix} shows a mixed view of the features with three channels colored in red, green and blue, respectively. 
The picture depicts a local centripetal pattern with the  center located near the upper left corner.
Fig.~\ref{fig:conv3} shows outputs of recurrent layer 3 (the last output which is used by the FC layer), from which we can observe that the hidden layers extract some local features such as the edge of fields and the direction of currents. Those feature maps with deep darkness correspond to other patterns.



\subsection{Evaluations}

We implemented two methods: one belongs to the control paradigm and we use the basic iLQR to compute the control inputs; the other one is the deep reinforcement learning (DRL) framework that employs the guided policy mechanism, where the policy is guided by (and combined with) the iLQR solving process~\cite{levine2013guided}.


\subsubsection{Artificial Disturbances}
We first investigate the method using artificially generated disturbances. We tested different vector fields including vortex, meandering, uniformly spinning (waving), and centripetal patterns. 
For different trials, we specify the robot with different start and goal locations, and the {\em goal reaching rate} is calculated by the number of successes divided by the total number of simulations. 
The results in Table~\ref{tab:goal reaching rate} show that within given time limits, both the iLQR and DRL methods lead to a good success rate, and particularly the DRL performs better in complex environments like the vortex field.
In contrast, the iLQR framework has a slightly better performance in relatively mild environments where the disturbance has slowly-changing dynamics, such as the meandering disturbance field. 
Then, we test the average time costs, as shown in Table~\ref{tab:average time}. 
The results reveal that the trials using iLQR tend to consume less time than those from the DRL method. 
This can be due to the ``idealized" artificial disturbances with known simple and accurate patterns, which can be precisely handled by the traditional control methodology.

\begin{table}[!hbp]
\begin{tabular}{c|c|c|c|c}
\hline
\hline
\tabincell{c}{Disturbance\\ pattern} & Method & \tabincell{c}{ Num of \\trials} & \tabincell{c}{Num of\\ success} & \tabincell{c}{Success rate} \\
\hline
\multirow{2}{*}{Vortex} & DRL    & 50 & 48 & 0.96 \\
\cline{2-5}            & iLQR & 50 & 46 & 0.92 \\
\hline
\multirow{2}{*}{Meander} & DRL    & 50 & 49 & 0.98 \\
\cline{2-5}           & iLQR & 50 & 50 & 1.00 \\
\hline
\multirow{2}{*}{Spin} & DRL    & 50 & 49 & 0.98 \\
\cline{2-5}              & iLQR & 50 & 48 & 0.96 \\
\hline
\multirow{2}{*}{Centripetal} & DRL    & 50 & 49 & 0.98 \\
\cline{2-5}                  & iLQR & 50 & 48 & 0.96 \\
\hline
\hline
\end{tabular}
\caption{Simulation with artificially generated disturbances}
\label{tab:goal reaching rate}
\end{table}

\begin{table}[!hbp]
\begin{tabular}{c|c|c|c}
\hline
\hline
Pattern & Method & Num of trials & Average time cost \\
\hline
\multirow{2}{*}{Vortex} & DRL    & 50 & 20.549 \\
\cline{2-4}            & iLQR & 50 & 14.811 \\
\hline
\multirow{2}{*}{Meander} & DRL    & 50 & 16.926 \\
\cline{2-4}           & iLQR & 50 & 15.367 \\
\hline
\multirow{2}{*}{Spin} & DRL    & 50 & 17.667 \\
\cline{2-4}              & iLQR & 50 & 17.803 \\
\hline
\multirow{2}{*}{Centripetal} & DRL    & 50 & 20.220 \\
\cline{2-4}                  & iLQR & 50 & 14.792 \\
\hline
\hline
\end{tabular}
\caption{Average time cost under artificial disturbances}
\label{tab:average time}
\end{table}

\begin{figure}[htbp]
  \centering
  \includegraphics[width=0.8\columnwidth]{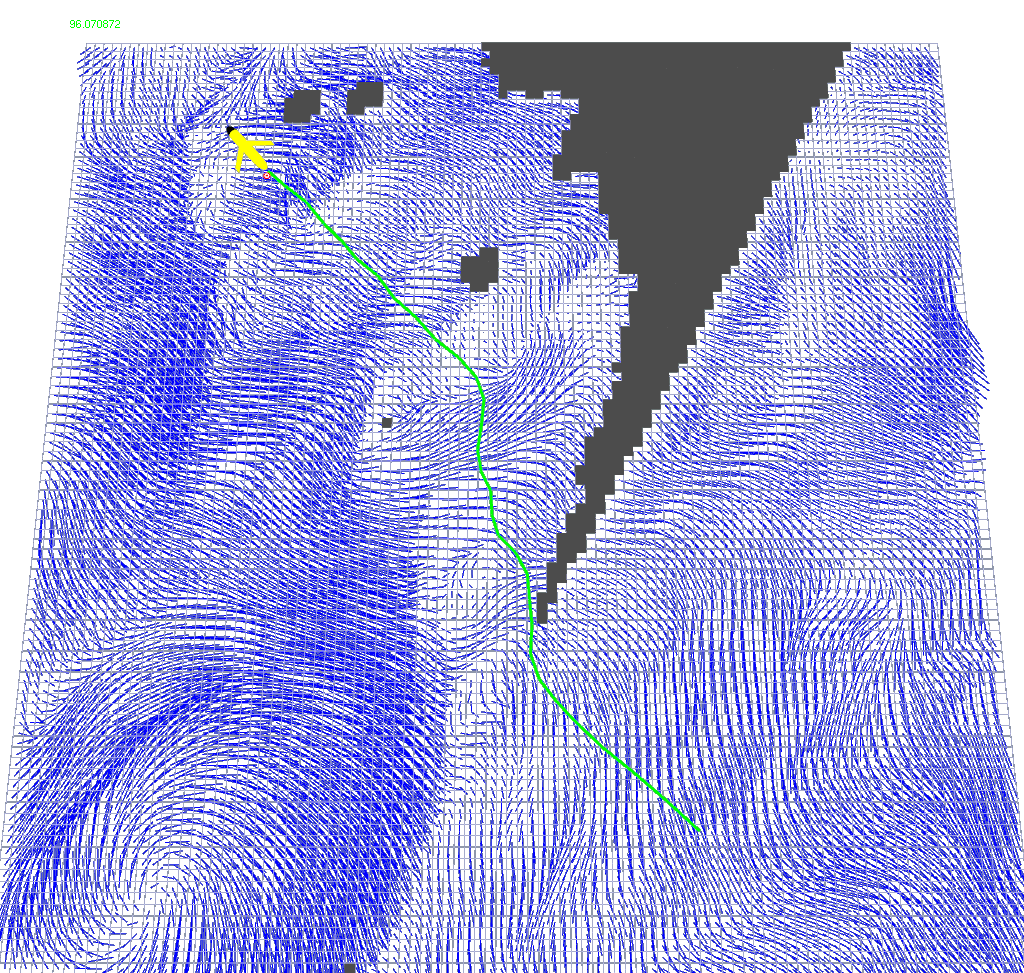}
  \caption{Demonstration of the ocean currents and a path of the robot}
  \label{fig:roms}
\end{figure}

\begin{figure}[htbp]
  \centering
  \subfigure[Time Cost]
  {
    \includegraphics[width=0.46\columnwidth]{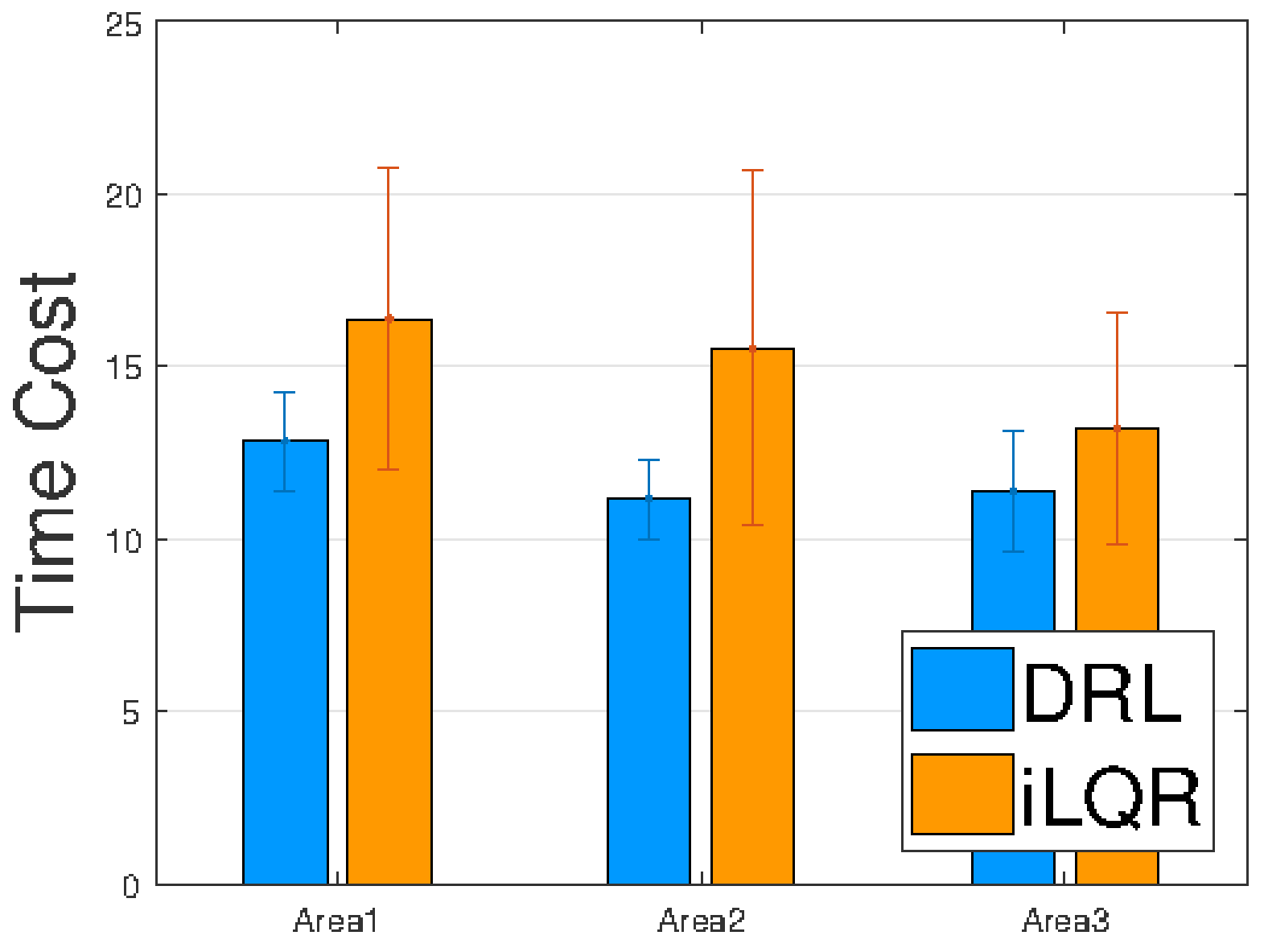}
    \label{fig:time_cost}
  }
  \subfigure[Step Cost]
  {
    \includegraphics[width=0.46\columnwidth]{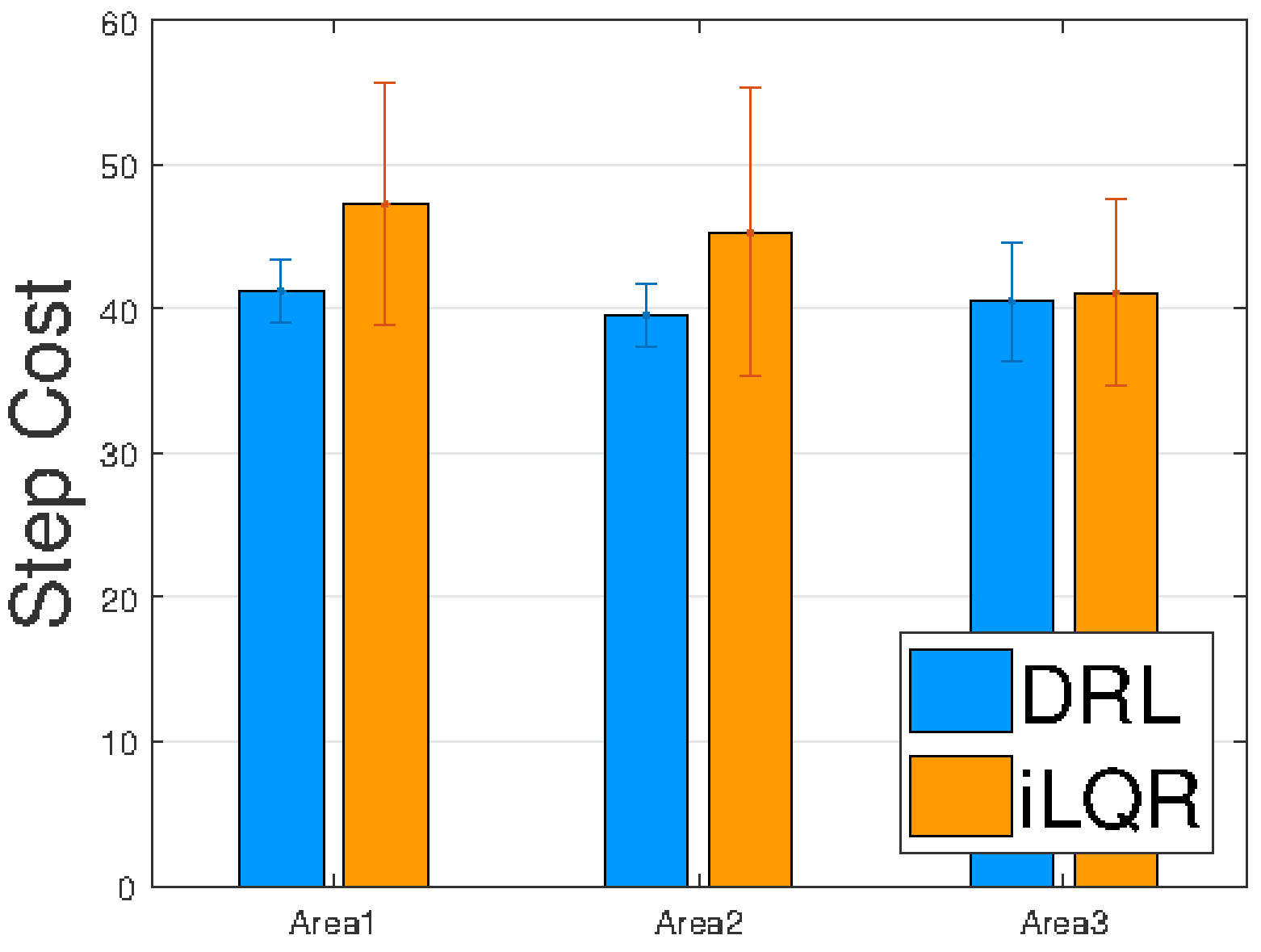}
    \label{fig:step_cost}
  }
  \caption{Average cost under ocean disturbances}
  \label{fig:avg_cost_fig}
\end{figure}

\begin{figure*}[t]
  \centering
  \subfigure[DRL]
  {
    \includegraphics[width=0.47\columnwidth]{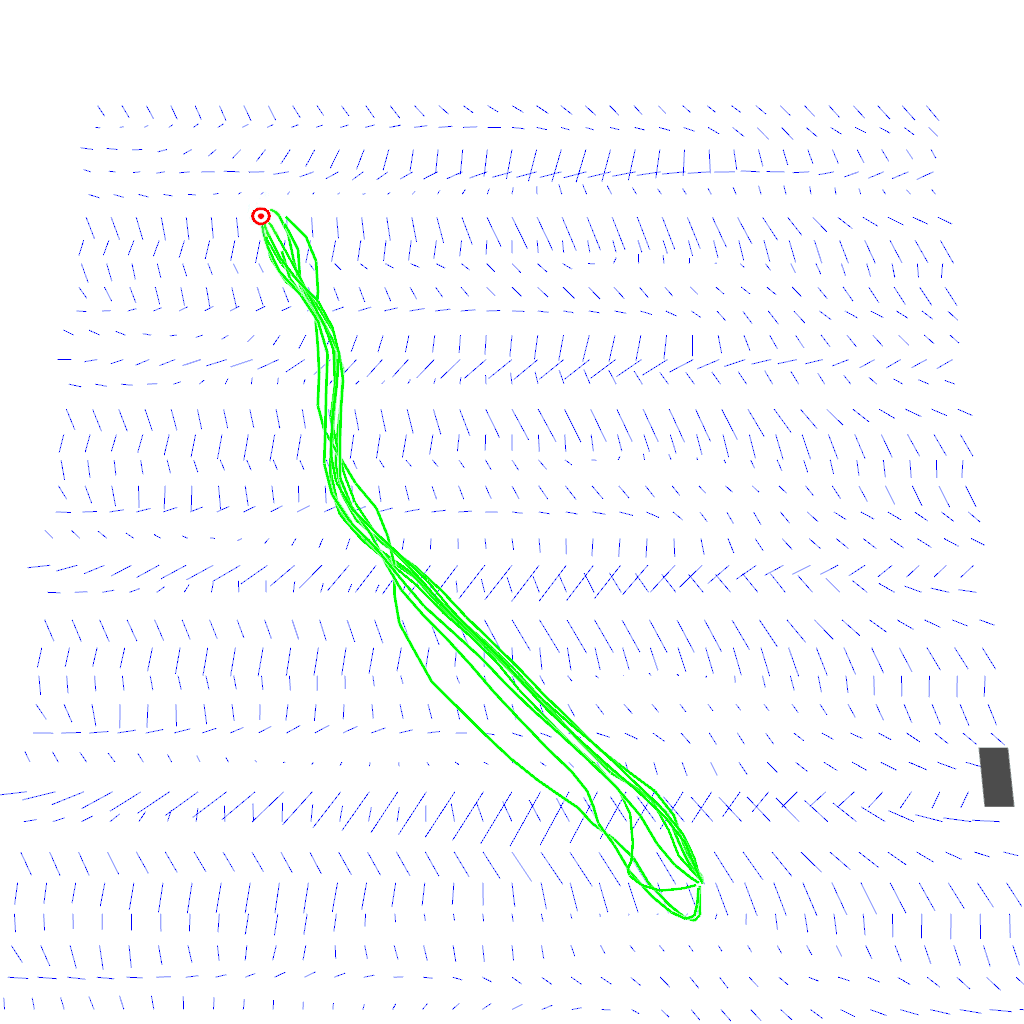}
    \label{fig:drl_syn}
  }
  \subfigure[]
  {
    \includegraphics[width=0.47\columnwidth]{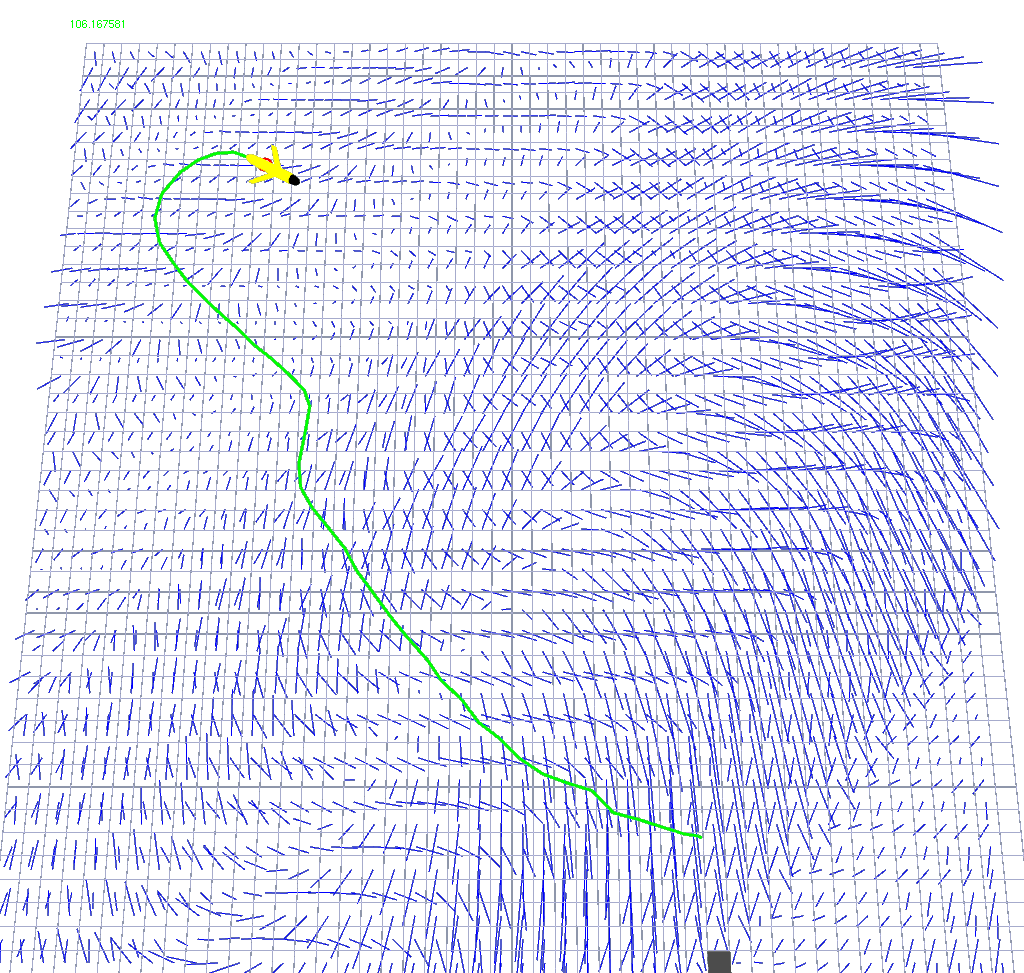}
    \label{fig:area3_1}
  }
  \subfigure[]
  {
    \includegraphics[width=0.47\columnwidth]{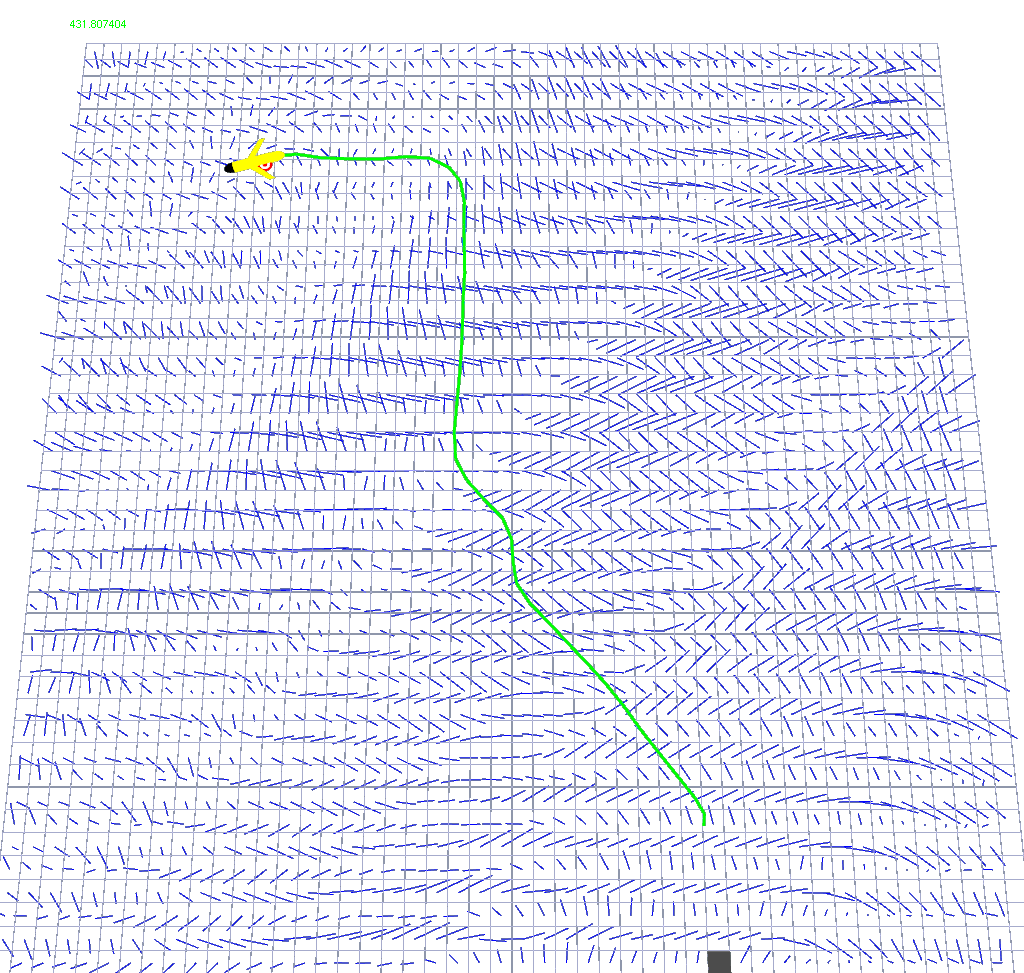}
    \label{fig:area3_2}
  }
  \subfigure[]
  {
    \includegraphics[width=0.47\columnwidth]{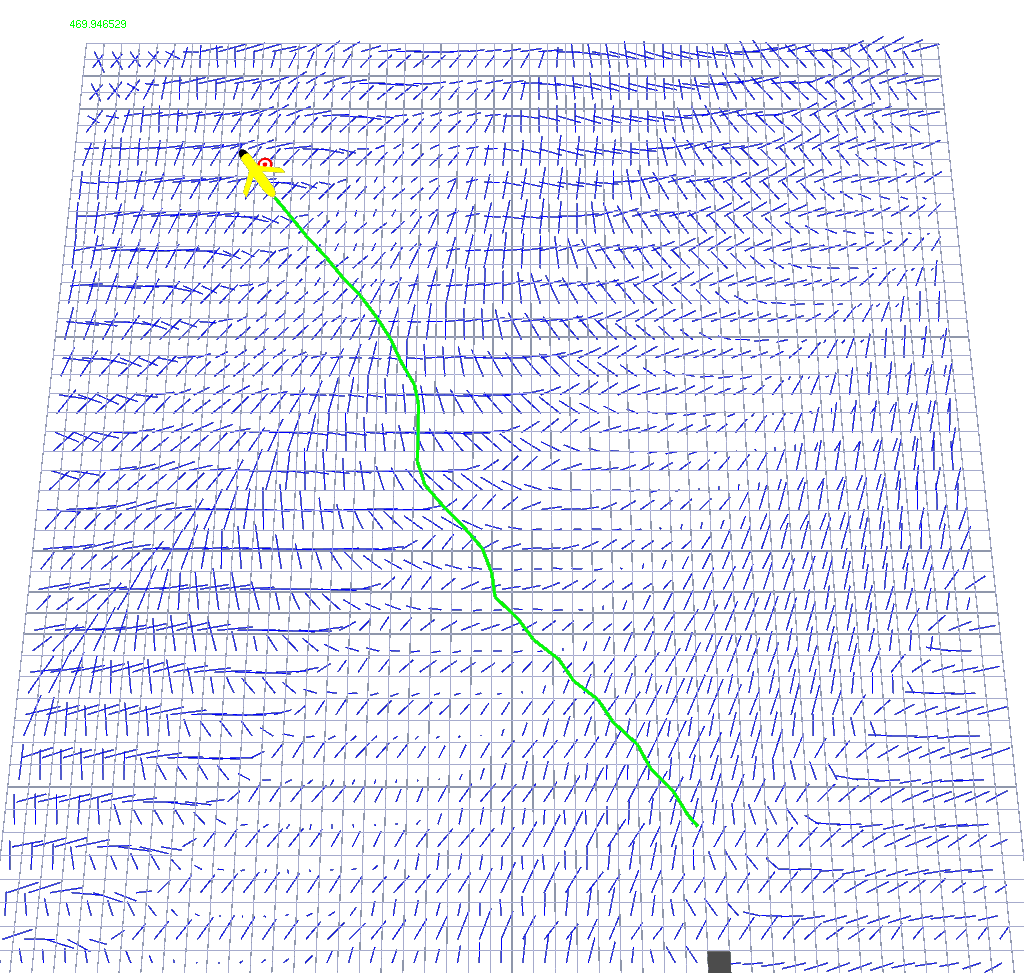}
    \label{fig:area3_3}
  }
  \caption{Illustrations of robot's motion trajectories. (a) Typical paths learned from DRL; (b) Examples of paths under different spatiotemporal disturbance patterns.}
  \label{fig:real_env}
\end{figure*}

\subsubsection{Ocean Data Disturbances}

In this part of evaluation, we use ocean current data obtained from the California Regional Ocean Modeling System (ROMS)~\cite{shchepetkin_regional_2005}.
The ocean data along the coast near Los Angeles is released every 6 hours and a window of 30 days of data is maintained and retrievable~\cite{yichao2017roms}. 
An example of ocean current surface can be visualized in Fig.~\ref{fig:roms}, which also demonstrates a robot's path from executing our training result.

Because the original ROMS ocean dataset covers a vast area and practically it requires several days for the robot to travel through the whole space, during which a lot of variation and uncertainty may occur. Thus, we opt to focus on smaller local areas and randomly cropped such sub-areas to evaluate our training results. 
Fig.~\ref{fig:drl_syn} illustrates a synthesized view of multiple trials by employing the DRL training result in a selected area of ocean data. Fig.~\ref{fig:area3_1} to \ref{fig:area3_3} illustrate different paths during training in various disturbance areas.

Similar to the evaluation process for the artificial disturbances, we looked into those aforementioned performances, i.e., the success rate and time cost, under the ocean disturbances. 
We also examined the total number of steps of each path, where each step represents the robot transiting from one grid to a new one (resolution of grid map is the same as the vector field map). 
Table.~\ref{tab:average time real} shows the results (robot speed does not scale to real map), from which we can see that the DRL takes less time (it saves around $20\%$ time on average comparing to iLQR) and fewer steps (by saving around $10\%$ on average). 

\begin{figure*}[t]
  \centering
  \subfigure[Example Area 1]
  {
    \includegraphics[width=0.6\columnwidth]{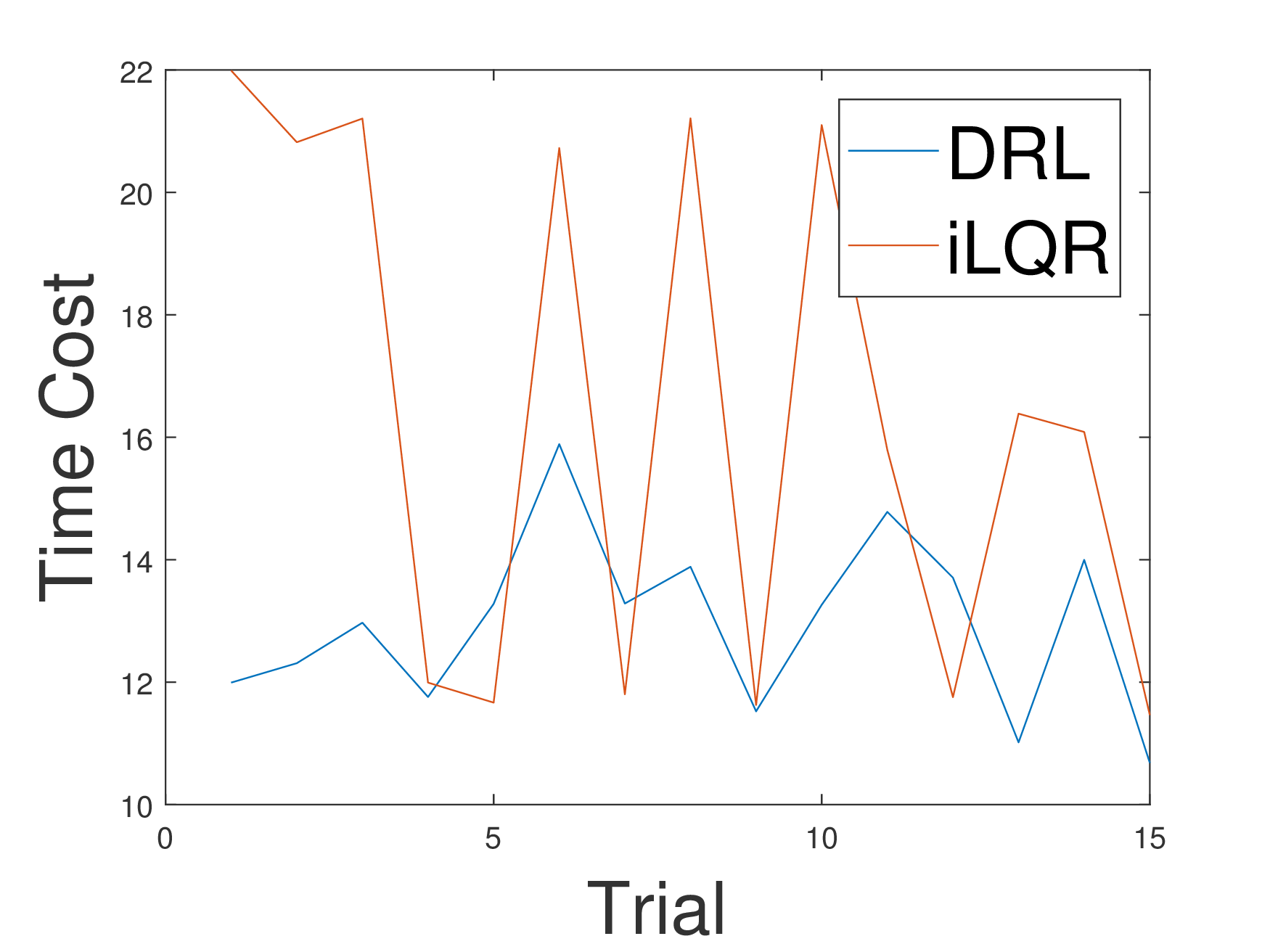}
    \label{fig:area1_tcost}
  }
  \subfigure[Example Area 2]
  {
    \includegraphics[width=0.6\columnwidth]{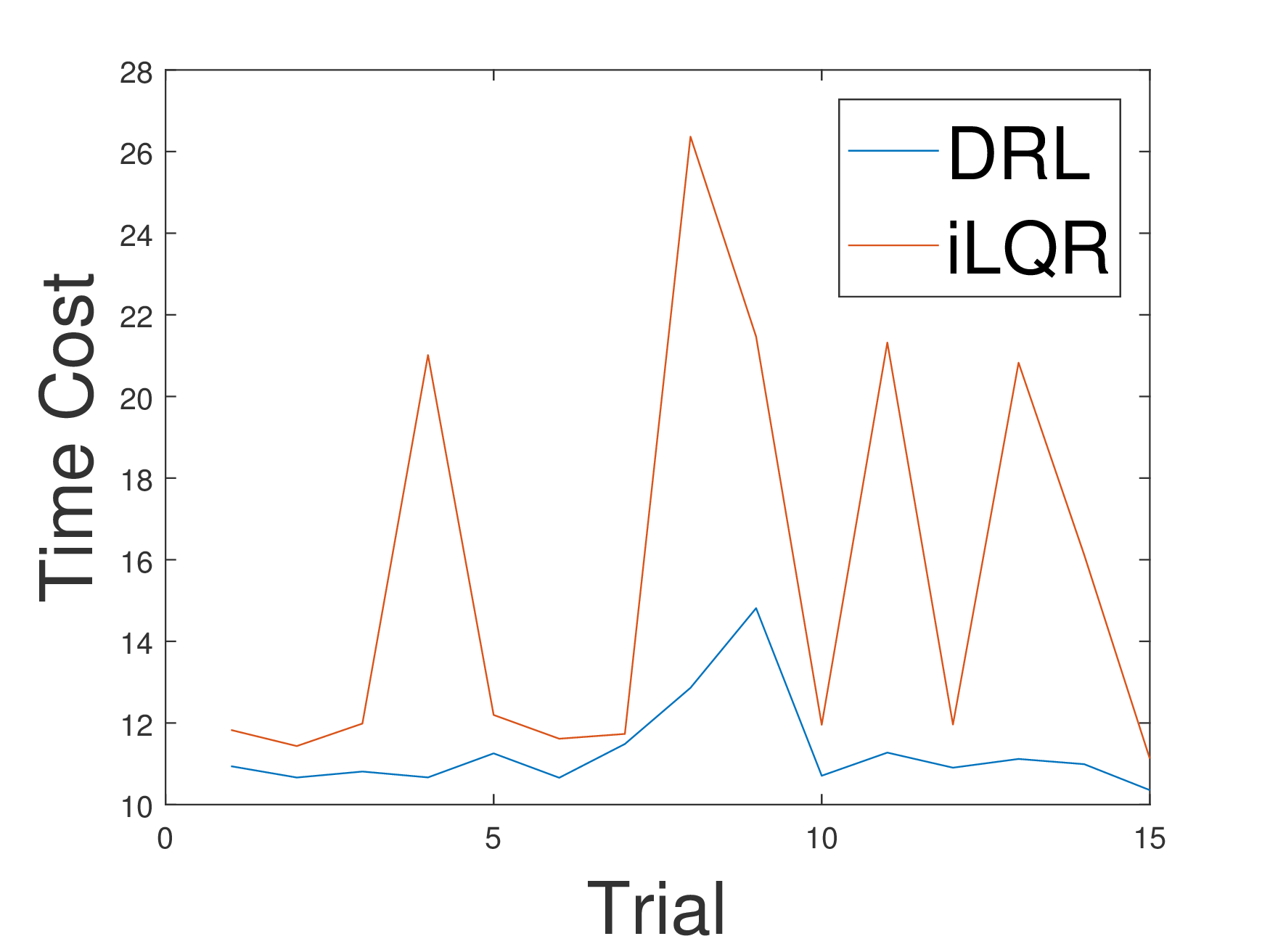}
    \label{fig:area2_tcost}
  }
  \subfigure[Example Area 3]
  {
    \includegraphics[width=0.6\columnwidth]{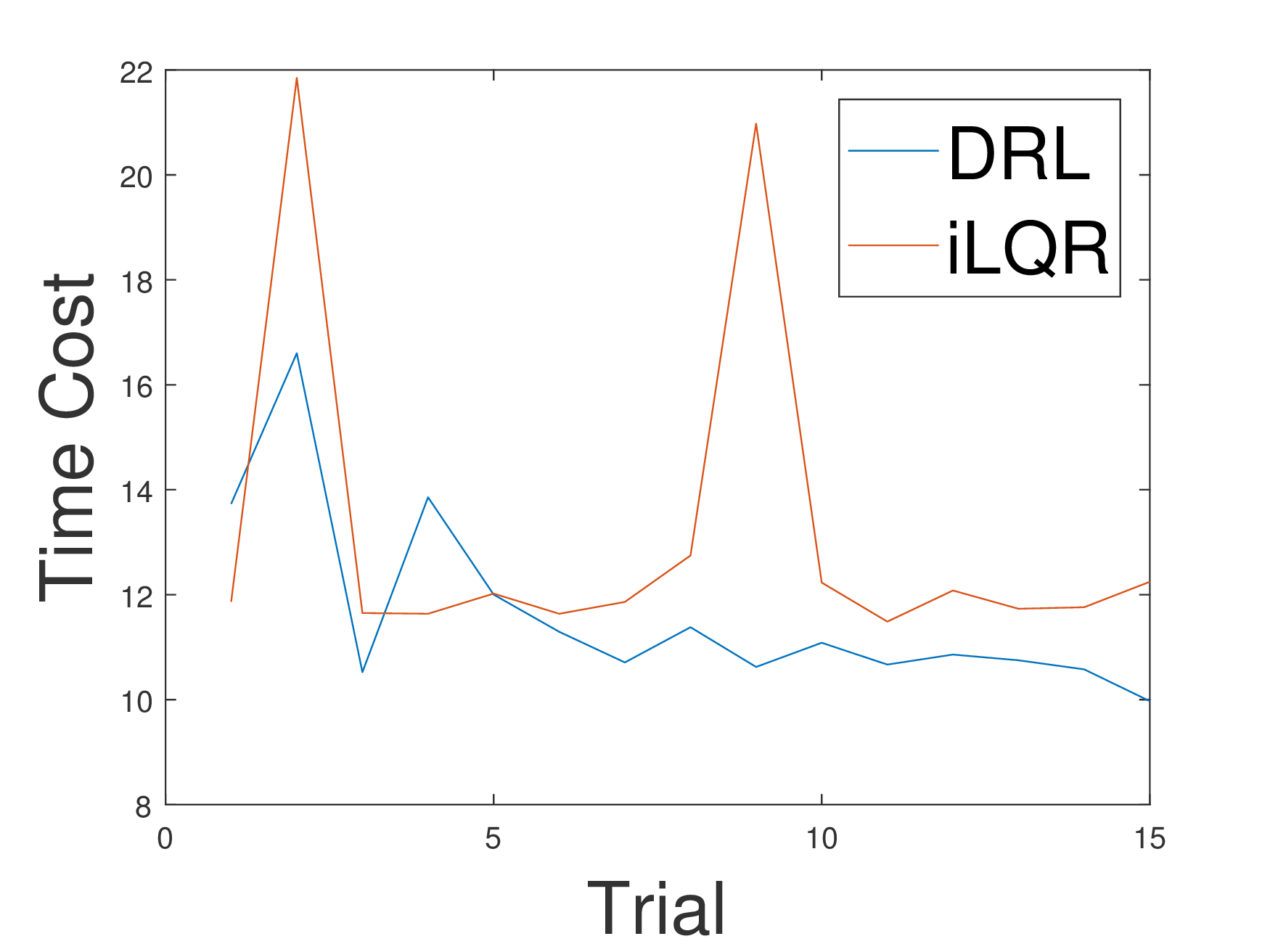}
    \label{fig:area3_tcost}
  }\\
  \subfigure[Example Area 1]
  {
    \includegraphics[width=0.6\columnwidth]{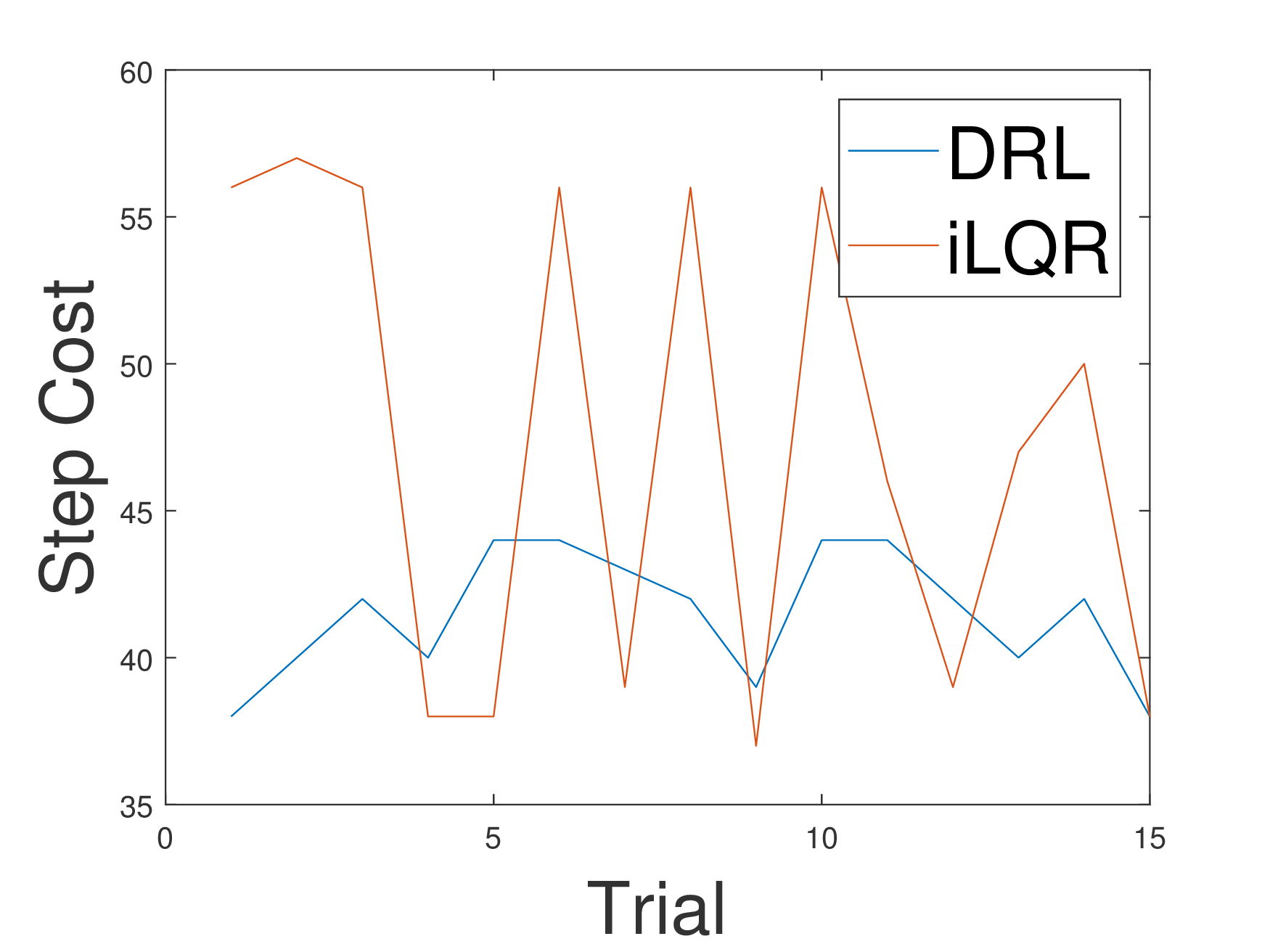}
    \label{fig:area1_scost}
  }
  \subfigure[Example Area 2]
  {
    \includegraphics[width=0.6\columnwidth]{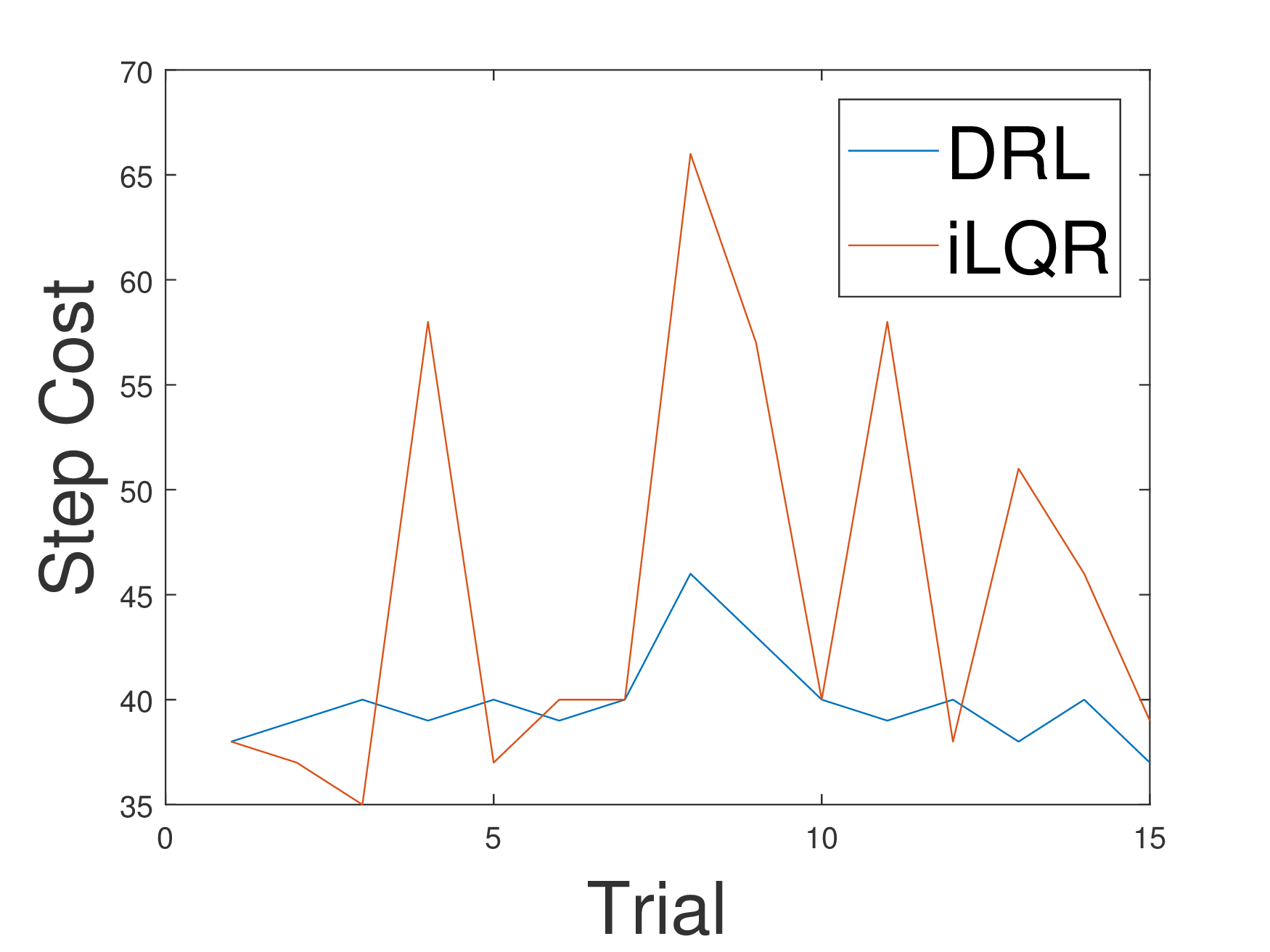}
    \label{fig:area2_scost}
  }
  \subfigure[Example Area 3]
  {
    \includegraphics[width=0.6\columnwidth]{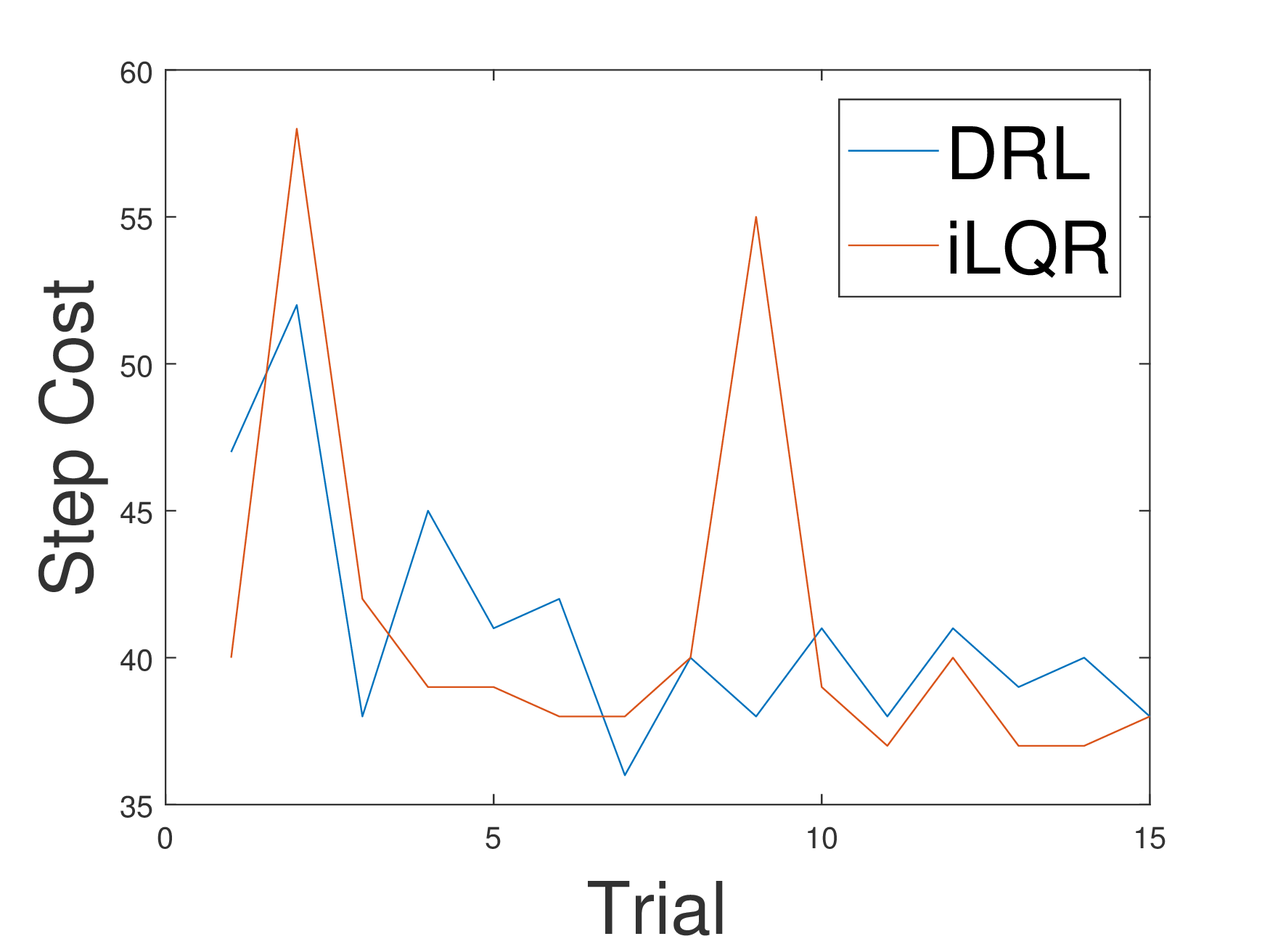}
    \label{fig:area3_scost}
  }
  \caption{Time and step costs under different ocean disturbances in three example areas}
  \label{fig:cost_fig}
\end{figure*}

Fig.~\ref{fig:avg_cost_fig} provides a better view for comparing the time and step costs, where statistical  variations can be closely examined. 
In detail, the standard derivation of DRL is much less than that of iLQR. 
Such characteristic can also be observed in Fig.~\ref{fig:cost_fig}, where those saw-tooth curves of iLQR imply dramatic changes among differing trials.

Our experimental evaluations indicate that the iLQR works well for those environments that can be well described and accurately formulated. 
In contrast, the DRL framework is particularly capable of handling complex and (partially) unstructured spatiotemporal environments that cannot be precisely modelled.


\begin{table}
\begin{tabular}{c|c|c|c|c|c}
\hline
\hline
Area & Method & \tabincell{c}{Num of \\trials} & Success rate & \tabincell{c}{Average\\time cost} & \tabincell{c}{Average\\step cost} \\
\hline
\multirow{2}{*}{Area 1} & DRL    & 15 & 1.00 & 12.833 & 41.30 \\
\cline{2-6}             & iLQR & 15 & 1.00 & 16.375 & 47.27 \\
\hline
\multirow{2}{*}{Area 2} & DRL    & 15 & 1.00 & 11.142 & 39.55 \\
\cline{2-6}             & iLQR & 15 & 1.00 & 15.530 & 45.33 \\
\hline
\multirow{2}{*}{Area 3} & DRL    & 15 & 1.00 & 11.383 & 40.50 \\
\cline{2-6}             & iLQR & 15 & 1.00 & 13.186 & 41.13 \\
\hline
\hline
\end{tabular}
\caption{Average time cost under ocean disturbances}
\label{tab:average time real}
\end{table}

\section{Conclusions}


In  this  paper we investigate applying the deep reinforcement learning framework for robotic learning and acting in partially-structured environments.
We use the scenario of marine vehicle decision-making under spatiotemporal disturbances to demonstrate and validate the framework. We show that the deep network well characterizes local features of varying disturbances.
By training the robot under artificial and real ocean disturbances, our simulation results indicate that the robot is able to successfully and efficiently act in complex and partially structured environments.


\appendix \label{sec:appendix}
This section will discuss the solutions to the dynamic model \eqref{eqn:time-varying-dynamics}.
We define the state of the robot as $s = \begin{pmatrix}x, y, \theta \end{pmatrix}^T$, and write the model dynamics in discrete time as
\begin{align}
s_{k+1} &= s_{k} + \begin{pmatrix}v_k \cos{\theta_k} + \omega_x (x_k,y_k)\\v_k \sin{\theta_k} + \omega_y (x_k,y_k)\\ u_k\end{pmatrix} \Delta t \nonumber\\
&:= f(s_k, u_k), \label{eqn:dynamics_basic}
\end{align}
where $\Delta t$ is the length of discrete time, and $k = 0, \ldots, N$. The initial state $s_0$ of the robot is known. We further assume that the total loss function $J_0$ to be minimized is of the form
\begin{equation}
J_{0} = \sum_{k=0}^{N-1} l(s_k, u_k) + l_f(s_N), \label{eqn:original_total_loss}
\end{equation}
where $l(\cdot, \cdot)$ and $l_f(\cdot)$ are in the quadratic form, i.e., \addtocounter{equation}{-1}
\begin{subequations}
\begin{align}
l_f(s_N)&=\frac{1}{2}\left(s_f-s_N\right)^TW_f\left(s_f-s_N\right),\label{eqn:original_total_loss_sub_a}\\
l(s_k, u_k)&=\frac{1}{2}(s_k^T W_p s_k + \rho u_k^2),\\ \intertext{where}
W_p &= W_f = \begin{bmatrix}1& 0& 0\\0& 1& 0\\0& 0& 0\end{bmatrix}, \label{eqn:original_total_loss_sub_c}
\end{align}
\end{subequations}
$\rho$ is a constant parameter, and $s_f = (x_f, y_f, \theta_f)^T$ is the position of final state.



It is typically difficult to search for solutions when the dynamic model is of a nonlinear form. To make the problem tractable, the iterative LQR (iLQR) algorithm is then employed \cite{tassa2014control}. The essential idea is based on the iteratively incremental improvement. Starting from a sequence of states $\{s_k^0\}$ and control variables $\{u_k^0\}$, we approximate the nonlinear dynamic model (\ref{eqn:dynamics_basic}) by a linear one, and then apply LQR algorithm (with the nonlinear dynamics (\ref{eqn:dynamics_basic}) in the backward pass) to obtain the next control sequence $\{u_k^1\}$ and state sequence $\{s_k^1\}$. By repeating this procedure until convergence, one can obtain the final state and control sequences.

Now assume the iLQR proceeds until the $(i+1)$-th iteration with the state sequence $\{u_k^i\}$ and control $\{s_k^i\}$. The dynamic model can be linearly approximated as
\begin{align}
\delta s_{k}^i &= \frac{\partial{f(s_k^i, u_k^i)}}{\partial{s}} \delta s_k^i + \frac{\partial{f(s_k^i, u_k^i)}}{\partial{u}} \delta u_k^i, \label{linearized_dynamics}
\end{align}
where $\delta s_{k}^i = s_{k}^{i+1} -s_{k}^{i}$, $\delta u_{k}^i = u_{k}^{i+1} - u_{k}^{i}$, and
\begin{align*}
\frac{\partial{f}}{\partial{s}} &=
\begin{bmatrix}1+\frac{\partial{\omega_x}}{\partial{x}}\Delta t& \frac{\partial{\omega_x}}{\partial{y}}\Delta t & -v\sin{\theta}\Delta t\\
\frac{\partial{\omega_y}}{\partial{x}}\Delta t& 1+\frac{\partial{\omega_y}}{\partial{y}}\Delta t & v\cos{\theta}\Delta t\\
0 & 0 & 1\end{bmatrix}, \\
\frac{\partial{f}}{\partial{u}} &= \begin{bmatrix}0\\0\\\Delta t\end{bmatrix}.
\end{align*} 
Note that the total loss function becomes $J_0(\{s_k^i+\delta s_k^i\}, \{u_k^i+\delta u_k^i\})$ where $\{s_k^i\}$ and $\{u_k^i\}$ are viewed as constants. Specifically, the total loss function can be expressed by the second-order expansion as
\begin{align}
J_0 &= \sum_{k=0}^{N-1} l(s_k^i+\delta s_k^i, u_k^i + \delta u_k^i) + l_f(s_N^i+\delta s_N^i),\nonumber \\
&= J(\{s_k^i\}, \{u_k^i\}) \nonumber \\
&+ \sum_{k=0}^{N-1}\left( l_{s_k}^T \delta s_k^i + \frac{1}{2}(\delta s_k^i)^T l_{s_k s_k} \delta s_k^i \right.\nonumber\\
&~\left.+ l_{u_k}^T \delta u_k^i + \frac{1}{2}(\delta u_k^i)^T l_{u_k u_k} \delta u_k^i\right) \nonumber \\
&+ (l_f)_{s_N}^T \delta s_N^i + \frac{1}{2}(\delta s_N^i)^T (l_f)_{s_Ns_N} \delta s_N^i, \label{eqn:approx_obj}
\end{align}
where
\begin{equation*}
\begin{aligned}[c]
l_{s}(s_k^i)&= W_p s_k^i,\\
 l_{u}(u_k^i) &= \rho~u_k^i, 
\end{aligned}
\qquad
\begin{aligned}[c]
l_{ss}(s_k^i) &= W_p,\\
l_{uu}(u_k^i) &= \rho.
\end{aligned}
\end{equation*}
Due to the form of loss functions \eqref{eqn:original_total_loss_sub_a}-\eqref{eqn:original_total_loss_sub_c}, there are no cross quadratic terms between $\delta s_k^i$ and $\delta u_k^i$.  
All the above equalities in (\ref{eqn:approx_obj}) are exact.
If, however, the loss function is nonlinear, the last equation with necessary cross terms is considered as a quadratic approximation. We then search for  $\{\delta s_k^i\}$ and $\{\delta u_k^i\}$ to minimize the $J_0$. The procedure is essentially the standard dynamic programming for the LQR problem \cite{liberzon2011calculus, Kat17}. During the backward iteration $k=N-1, \ldots, 0$, the feedback gain matrix $\mathbb{K}_k$ is computed to get the optimal control update $(\delta u_k^i)^*$,
\begin{align*}
(\delta u_k^i)^* &= \mathbb{K}_k\delta s_k^i + \mathbb{k}_k,
\end{align*}
where
\begin{equation*}
\begin{aligned}[c]
\mathbb{K}_k &= -\mathbb{Q}_{u_ku_k}^{-1}\mathbb{Q}_{u_ks_k},
\end{aligned}
~\mbox{and}\quad
\begin{aligned}[c]
\mathbb{k}_k &= -\mathbb{Q}_{u_ku_k}^{-1}\mathbb{q}_{u_k}. 
\end{aligned}
\end{equation*}
The related matrices in above equations can be obtained through the following functions
\begin{equation*}
\begin{aligned}[c]
\mathbb{Q}_k &= \mathbb{L}_k + \mathbb{F}_k^T \mathbb{M}_{k+1} \mathbb{F}_k = \begin{bmatrix}\mathbb{Q}_{s_ks_k}, & \mathbb{Q}_{s_ku_k}\\\mathbb{Q}_{u_ks_k},& \mathbb{Q}_{u_ku_k}\end{bmatrix},\\
\mathbb{q}_k &= \mathbb{l}_k + \mathbb{F}_k^T \mathbb{m}_{k+1}=\begin{bmatrix}\mathbb{q}_{s_k}\\ \mathbb{q}_{u_k}\end{bmatrix}, 
\end{aligned}
\end{equation*}
and

\begin{equation*}
\begin{aligned}[c]
\mathbb{M}_k &= \mathbb{Q}_{s_ks_k} - \mathbb{K}_k^T \mathbb{Q}_{u_ku_k}\mathbb{K}_k,\\
\mathbb{m}_k &= \mathbb{q}_{s_k} - \mathbb{K}_k^T \mathbb{Q}_{u_ku_k}\mathbb{k}_k,\\
\end{aligned}
\end{equation*}

where we denote the joint dynamic matrix $\mathbb{F}_k=\left[\frac{\partial{f}}{\partial{s}}, \frac{\partial{f}}{\partial{u}}\right]$, and value matrices
\begin{equation*}
\begin{aligned}[c]
\mathbb{L}_k &=  \begin{bmatrix}l_{ss}(s_k^i), & 0\\0,& l_{uu}(u_k^i)\end{bmatrix},
\end{aligned}
~\mbox{and}\quad
\begin{aligned}[c]
\mathbb{l}_k &=\begin{bmatrix}l_{s}(s_k^i)\\ l_{u}(u_k^i)\end{bmatrix}. 
\end{aligned}
\end{equation*}
After the backward iteration, there is a forward iteration to compute $u_k^{i+1}$ by 
\begin{equation*}
u_k^{i+1} = u_k^i + \mathbb{K}_k( s_k^{i+1} - s_k^{i}) + \mathbb{k}_k,
\end{equation*}
and to update the $s_k^{i+1}$ by the dynamic model \eqref{eqn:dynamics_basic} with the initial state.  
This completes the $(i+1)$-iteration in the iLQR framework.

 { 
 \bibliographystyle{abbrv}
 \bibliography{reference}
 }

\end{document}